# NumtaDB - Assembled Bengali Handwritten Digits

Samiul Alam[1], Tahsin Reasat [1], Rashed Mohammad Doha [2], Ahmed Imtiaz Humayun [1],
[1]*Department of EEE, Bangladesh University of Engineering & Technology*, Dhaka, Bangladesh
[2]*Department of Mechanical Eng, Bangladesh University of Engineering & Technology*, Dhaka, Bangladesh

*Abstract*—To benchmark Bengali digit recognition algorithms, a large publicly available dataset is required which is free from biases originating from geographical location, gender, and age. With this aim in mind, *NumtaDB*, a dataset consisting of more than 85,000 images of hand-written Bengali digits, has been assembled. This paper documents the collection and curation process of numerals along with the salient statistics of the dataset.

## I. INTRODUCTION

Hand written digit recognition (HWDR) is a classic problem in the area of computer vision. There are various practial application of an HWDR system, for example, ZIP code recognition [1] and reading bank checks [2]. Design of an accurate HWDR system requires a dataset which is collected from multiple contributors to count for the variation in individual writing style. The initial research in this sector focused on the recognition of English digits and it produced the iconic *MNIST* database [2]. Gradually, research in HWDR spread in other language and researchers have created hand written digit dataset in French [3], Farsi [4]–[6], Urdu [7], Chinese [8] , Kannada [9], Oriya, Tamil, Telegu [10], Devnagari [11], [12] etc.

Bengali is the official language of Bangladesh and second most widely spoken language of India, behind Hindi. It has approximately 189 million native and 208 million total speakers worldwide [13]. Bengali is the seventh most spoken native language in the world by population [14]. Although Bengali is a thousand year old language, research in handwritten characters did not initiate until the mid nineties [15]. The first publicly available database on handwritten numerals was released by Indian Statistical Institute (ISI) [16]. Although 23392 images were included in this dataset only 500 images are hosted on the website. Later, the Center for Microprocessor Application for Training Education and Research (CMATER) group released the *CMATERDB* dataset [17]. But the dataset is not publicly available anymore. Recently, two datasets have been produced and made available by the researchers at Shahjalal University of Science and Technology (SUST), University of Liberal Arts (ULAB) and University of Asia Pacific (UAP) [18], [19]. The SUST-Bangla Handwritten Numeral Database (*SUST-BHND*) has 101065 sample image. However, majority of the data contributors are from Sylhet and the gender distribution is not reported. The *BanglaLekha-Isolated* database produced by ULAB and UAP contains Bengali characters along with 19,748 numerals. The contributors belong to educational institutes in Dhaka and Comilla. Unlike the other databases, it includes samples from children.

The current difficulty faced by the researchers in HWDR is the lack of large publicly available database which is unbiased in terms of geographic location, age, and gender. Hence, for the proper benchmarking of Bengali HWDR algorithms, there is a profound need for a standardized open sourced dataset which is publicly available for researchers. To mitigate this challenge, we have assembled numerals from several sources and combined them together to form an open sourced dataset *NumtaDB*. Most of the data were collected from students of the public universities (funded by the government while managed as self-governed organizations) in Dhaka. Since it is well known that the students in these universities come from all over Bangladesh, we have implicitly made sure representatives from most of the regions are present in the samples. To account for the presence of samples from children we have incorporated the numerals from *BanglaLekha-Isolated* with permission from the owners. The images have been checked under strict guidelines of legibility so that there is minimum noise in the samples. This paper is organized in the following way. Section II describes the different sources of data and contributor information, Section III describes the digit extraction procedure from the scanned images, Section IV describes the legibility criterion of digits and evaluation procedure, V presents the dataset statistics after it has been split into train and test set, and finally the paper is concluded through Section VI.

## II. DATA SOURCES

The dataset is a combination of six datasets that were gathered from different sources and at different times. Information regarding contributors and image collection is summarized in Table I and information regarding image collection and curation is summarized in Table II.

### A. Bengali Handwritten Digits Database (BHDDB)

BHDDB dataset was collected from students in the Department of Computer Science and Engineering of Bangladesh University of Engineering and Technology. The students were given a form to write down the numerals. The forms had regular grid pattern in which the numerals were inscribed. The forms were scanned in color. The forms had a marker on each of its four corners which were used to align the image borders with the grid lines. The digit extraction procedure is described in Section III.

TABLE I
CONTRIBUTOR INFORMATION

| Dataset Name | Date of collection | Data Source | Age Range | Male/Female Ratio (%) | No. of contributors | No. of digits per contributor | Frequency of digits per contributor |
|---|---|---|---|---|---|---|---|
| Bengali Handwritten Digits Database (BHDDB) | 17.12.17 | BUET, Schools and Colleges in Dhaka | 6-20 | 65/35 | 260 | 90 | 9 |
| BUET101 Database (B101DB) | 15.12.17 | BUET | 18-24 | 70/30 | 45 | 10 | 1 |
| OngkoDB | 17.12.17 | Department of CSE, BUET | 20-24 | 70/30 | 289 | 100 | 10 |
| DUISRT | Dec '18 | Dhaka | 20-24 | 51.3/48.7 | 145 | 90 | 9 |
| Bangla Lekha-Isolated | Sept'16 to Nov'16 | Dhaka and Comilla | 6-28 | 59.4/40.6 | 1988 | 10 | 1 |
| UIUDB | 15.12.17 | United International University, Mentors' | 18-25 | 75/25 | 15 | 40 | 4 |

TABLE II
IMAGE COLLECTION AND CURATION INFORMATION

| Dataset Name | Medium (Scanner/ Paint) | Formats | Total Number of Digits | Number of Digits Removed | Dimension of images |
|---|---|---|---|---|---|
| Bengali Handwritten Digits Database (BHDDB) | Data collected on forms and digitized with a scanner | PNG, 24 BIT COLOR | 23400 | 209 | 180x180 (Fixed) |
| BUET101 Database (B101DB) | Data Collected on paper and scanned at 600 dpi | PNG, 24 BIT COLOR | 435 | 7 | width: 94 to 110 height: 90 to 110 |
| OngkoDB | Data collected on forms and digitized with a scanner | PNG, 8 BIT GRAY-SCALE | 28900 | 321 | 180x180 (Fixed) |
| DUISRT | Scanned from paper forms | PNG 24 BIT COLOR | 13133 | 277 | 180x180 (Fixed) |
| Bangla Lekha-Isolated | Scanned from paper forms | PNG, 8 BIT BINARY | 20319 | 572 | width: 29 to 267 height: 266 to 180 |
| UIUDB | Scanned from paper, MS Paint, Cellphone Camera | 319 JPG, 257 PNG. | 576 | 81 | width: 63 to 879 height: 73 to 765 |

TABLE III
DATASET SUMMARY

| Original Name | Codename | Train-Test Split | Total Digits (Training) | Total Digits (Testing) | Total Digits (Combined) |
|---|---|---|---|---|---|
| BHDDB | a | 85%-15% | 19702 | 3489 | 23191 |
| B101DB | b | 85%-15% | 359 | 69 | 428 |
| OngkoDB | c | 85%-15% | 24298 | 4381 | 28679 |
| DUISRT | d | 85%-15% | 10908 | 1948 | 12856 |
| Bangla Lekha Isolated | e | 85%-15% | 16777 | 2970 | 19747 |
| UIUDB | f | 0%-100% | | 495 | 495 |
| Total | | | 72044 | 13552 | 85596 |

### B. BUET101 Database (B101DB)

The participants of this dataset wrote the digits on papers which were scanned in color at 600 dots per inch (dpi). The digits were than manually cropped and labeled.

### C. OngkoDB

OngkoDB was collected from a group of students from the Department of Computer Science and Engineering of Bangladesh University of Engineering and Technology. They filled up forms which did not have markers on the corners. The forms were scanned in gray-scale. Since the forms had no markers, a different extraction approach was taken. The images of the digits were extracted by first re-orienting feature points of SURF (Speeded Up Robust Feature) of the original image to a reference image and then extracting all images of the digits. The automated extraction procedure was not fully accurate and the dataset went through rigorous pruning (Details in Section III).

### D. ISRTHDB

ISRTHDB was collected from students in Institute of Statistical Research and Training, Dhaka University. The collection process and evaluation followed here was done in the same format as BHDDB dataset. This dataset was collected after BHDDB and had strong collaboration with people involved in the former. As such, the raw data of this dataset is much cleaner than its predecessor.

### E. BanglaLekha-Isolated Numerals

BanglaLekha-Isolated [19] dataset contains Bangla handwritten numerals, basic characters and compound characters. The data was collected from literate native bengali speakers from Dhaka and Comilla. The digits in this dataset contained erroneous labels and outliers which were cleaned and included in our dataset. The Banglalekha-Isolated dataset were released as preprocessed binary images. According to the authors, the following preprocessing steps were taken:

- Foreground and background were inverted so that images have a black background with the letter drawn in white.
- Noise removal was attempted by using a median filter.
- An edge thickening filter was applied.
- Images were resized to be square in shape with appropriate padding applied to preserve the aspect ratio of the drawn character.

*F. UIUDB*

UIUDB dataset was collected by students of United International University from scanned documents, windows paint images and cell phone camera photos. Due to the nature in which the data was gathered, this dataset is the hardest to train on and we have left it only in test set.

### III. EXTRACTION PROCEDURES

The large majority of the data excluding Banglalekha Dataset was extracted following one of two algorithms depending on whether markers were present in the forms. The resulting extracted data and the corresponding problems in each algorithm have been illustrated in the following sub sections.

*A. Marker based alignment and Grid Detection*

In case of BHDDB and ISRTDB, the raw scans had square markers. The forms had four markers placed at four edges of the region of interest which contain a rectangular grid like table. Digits were hand written inside the grids. The raw images are denoted by $I^R$ such that $I^R \in \mathbb{R}^{W,H}$. Here, $W$ and $H$ is the dimension of the image. The images were transformed into binary images $I^B$ and segmented into blobs (a connected area in an image). The set of blobs is denoted as $\mathcal{B}$. For each blob $b \in \mathcal{B}$, perimeter $P_b$, Area $A_b$, centroid $CE_b$, and bounding box $(W_b, H_b)$ were measured. Two properties circularity $C_b$ and extent $E_b$ were then defined as follows:

$$C_b = \frac{P_b^2}{4\pi A_b} \quad (1)$$

$$E_b = \frac{A_b}{W_b H_b} \quad (2)$$

The possible centroids of the markers, were determined by the segmented areas that satisfied the conditions defined as:

$$1.1 \leq C_b \leq 1.6 \quad (3)$$
$$E_b \geq 0.5 \quad (4)$$

The set of centroids that fulfill the condition is denoted by $\mathcal{V}$. If three marker centroids could be determined, the image was transformed and cropped to a rectangle whose vertices lies on those centroids. If more than three were found, then the four centroids that formed the rectangle with height and width closest to a reference rectangle were picked and the image was transformed accordingly. The reference rectangle was defined using the dimensions (height $H$ and width $W$) of the raw image. Then the horizontal and vertical lines of the grid would be in alignment with the axes. The aligned images ($I^A$) were then summed in each direction separately which outputs two arrays. These arrays would have strong peaks along grid lines (Fig. 1). So by using peak detection, all intersection points of grid lines were determined.

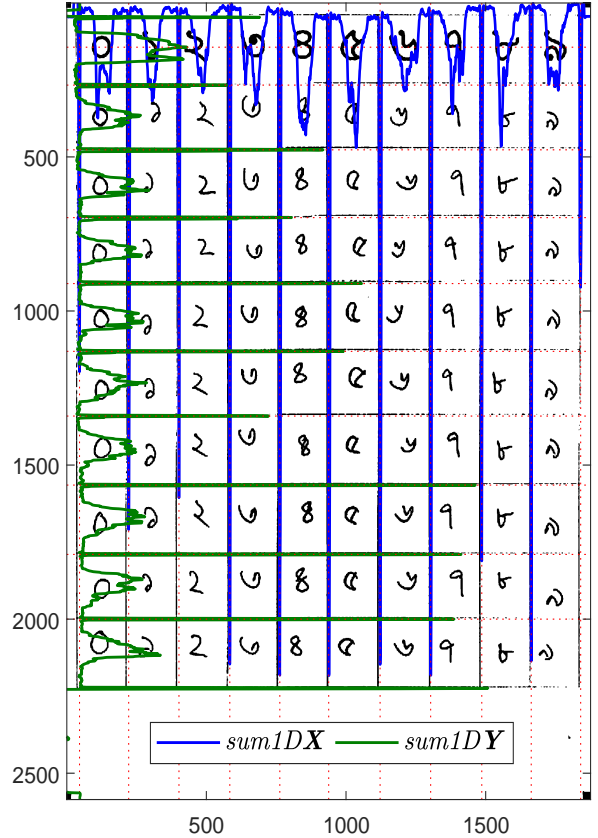

Fig. 1. Summation of the aligned image along each axis creates two one dimensional signals with distinct peaks (shown in green and blue) along the grid lines.

By using the points, the crops of the hand written digits ($d$) were extracted and included into the set of extracted digits denoted by $\mathcal{D}$. The digit extraction algorithm is illustrated in Algorithm 1. Since there was no margin between each grid box, some of the images had extensions from adjacent boxes intruding in their box (Fig. 4(g)). These were manually sorted out later.

*B. Markerless Alignment with SURF and Square Detection*

In case of OngkoDB, there was no marker. So an empty form was used as reference image which was perfectly aligned and using Speeded Up Robust Features, we realigned all the scanned image to that reference image (Fig. 2). Then the

**Algorithm 1:** Digit extraction from marker based forms

**Input** : Raw Scan of image, $\{I^R_{i,j}\}_{i,j=1}^{W,H}$
**Output:** Cropped set of digits, $\mathcal{D}$
**Initialize:** $\mathcal{V} \leftarrow \{\phi\}, \mathcal{D} \leftarrow \{\phi\}$

*Pre-Processing:*
$I^B \leftarrow$ Binarize($I^R$)
$I^B \leftarrow$ MedianFilter($I^B$, window = $5 \times 5$)
$\mathcal{B} \leftarrow$ BlobDetector($I^B$)
**if** $|\mathcal{B}| \leq 2$ **then**
 $I^B \leftarrow$ MedianFilter($I^B$, window = $15 \times 15$)
 $\mathcal{B} \leftarrow$ BlobDetector($I^B$)
**end**

*Determine rectangle vertices:*
**for** $\forall b \in \mathcal{B}$ **do**
 $CE_b, P_b, A_b, W_b, H_b \leftarrow$ BlobProperties($b$)
 $C_b \leftarrow \dfrac{P_b^2}{4\pi A_b}$
 $E_b \leftarrow \dfrac{A_b}{W_b H_b}$
 **if** $1.1 \leq C_b \leq 1.6 \cap E_b \geq 0.5$ **then**
  $\mathcal{V} \leftarrow \mathcal{V} \cup CE_b$
 **end**
**end**

*Align image and crop digits:*
$rectRef \leftarrow$ Rectangle($(0,0),W,H$)
$I^A \leftarrow$ GeometricTransfor($rectRef, \mathcal{V}, I^B$)
$sum1dX \leftarrow \sum_i \{I^A_{i,j}\}$
$sum1dY \leftarrow \sum_j \{I^A_{i,j}\}$
$peakX \leftarrow$ peakDetect($sum1dX$)
$peakY \leftarrow$ peakDetect($sum1dX$)
**for** $i = 1$ *to* $|peakX| - 1$ **do**
 **for** $j = 1$ *to* $|peakY| - 1$ **do**
  $m \leftarrow peakX_i$
  $n \leftarrow peakY_j$
  $m' \leftarrow peakX_{i+1}$
  $n' \leftarrow peakY_{j+1}$
  $d \leftarrow \{I^A_{i,j}\}_{i=m,l=n}^{m+m',n+n'}$
  $\mathcal{D} \leftarrow \mathcal{D} \cup d$
 **end**
**end**
**Return** $\mathcal{D}$

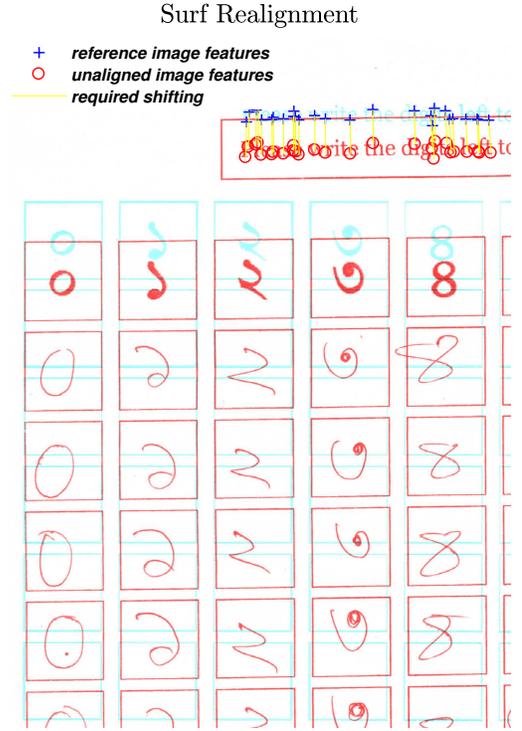

Fig. 2. Upper right portion of a marker less form is shown. The unaligned digit boxes (red) are aligned to the reference digit boxes (cyan) using SURF.

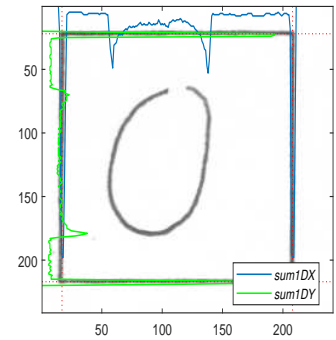

Fig. 3. Summing the pixels along X and Y axes creates peaks in the border region.

centroids of the squares in the reference image were used to cut out digits boxes from the realigned image. The width and height of the cut out were slightly larger than the box containing the digits so that the cropped image had both the hand written digit and the bounding box. Then the pixels were summed up in each direction separately which produced two arrays. These arrays would have strong peaks along bounding box lines (Fig. 3). By using peak detection the borders were detected and the digit were extracted.

## IV. LEGIBILITY CRITERION

Each of the datasets were examined under the same criteria to evaluate legibility. The following steps illustrate the procedure.

- All extracted images were grouped into ten separate folders corresponding to their digits. Then all images in each folder were examined by at least two people separately. During this stage most of the digits removed were improperly extracted or were blank or contained other numbers.
- The filtered dataset was rearranged into two separate folders corresponding to even or odd digits. They were re-examined for legibility. This ensured minimum priori knowledge was available during examination. Digits which were visually ambiguous were discarded in this step.

- The entire data was then merged into one single folder and were skimmed one final time to ensure that the data was free of outliers.

Some examples of discarded images are shown in Fig. 4.

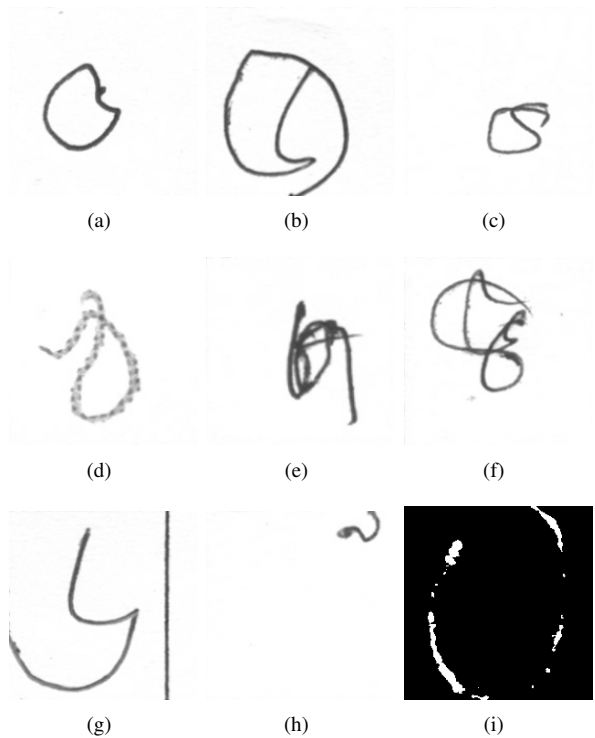

Fig. 4. Examples of handwritten digits discarded during manual checking. 4(a),4(b), and 4(c) were eliminated as they were not properly identifiable as a digit five. 4(d) could not be identifiable as any Bengali digit. 4(e) and 4(f) were eliminated for overwriting. 4(g) and 4(h) were discarded as the digits extended out of the bounding box. Also, 4(g) contains part of the grid line. 4(i) was discarded for having incomplete pen stroke

## V. Dataset Statistics

After all the illegible digits were pruned, the dataset was divided into training and testing set. The train-test split was done in a 85%-15% ratio for all the datasets except UIUDB. The splitting was done so that all digits in the training set were written by people who did not contribute in the testing set. Also the number of images per digit was kept approximately equal. The entire UIUDB dataset is kept in the test set. The statistics of both training and testing sets are given in the Table III.

## VI. Conclusion

In this paper, we have assembled data from isolated datasets gathered from over 2700 contributors with a view to maintaining variety within the data. The datasets were checked rigorously following a rigid methodology to ensure legibility of labels. The combined dataset, therefore, has accurate ground truths while maintaining a wide variety in terms of age groups, gender and location. The training set can be dowloaded from www.bengali.ai.

## VII. Acknowledgment

We are deeply grateful to the researchers behind *BanglaLekha-Isolated* for providing us their dataset. We would also like to thank the teams BUET Broncos, Shongborton, BUET Backpropers and UIU Kingkortobbobimurgh who participated in National Robotech Festival 2017 and contributed in our datasets. Lastly, we thank Tanvir Muhammed and his team from Institute of Statistical Research and Training, Dhaka University for their invaluable contributions in gathering data for the DUISRT dataset.